\documentclass[times,12pt,authoryear]{elsarticle}

\usepackage{hyperref}

\usepackage{amssymb}
\usepackage{gensymb}
\usepackage{xcolor}
\usepackage{amsmath}
\usepackage{float}
\usepackage[ruled,vlined,linesnumbered]{algorithm2e}

\begin{document}

\begin{frontmatter}



\title{ERA*: Enhanced Relaxed A* algorithm for Solving the Shortest Path Problem in Regular Grid Maps}


\author{Adel Ammar}

\affiliation{organization={RIOTU Lab, Prince Sultan UNiversity},
            addressline={Rafha Street}, 
            city={Riyadh},
            postcode={11586}, 
            country={Saudi Arabia}}

\begin{abstract}
This paper introduces a novel algorithm for solving the point-to-point shortest path problem in a static regular 8-neighbor connectivity (G8) grid. This algorithm can be seen as a generalization of Hadlock algorithm to G8 grids, and is shown to be theoretically equivalent to the relaxed $A^*$ ($RA^*$) algorithm in terms of the provided solution's path length, but with substantial time and memory savings, due to a completely different computation strategy, based on defining a set of lookup matrices. Through an experimental study on grid maps of various types and sizes (1290 runs on 43 maps), it is proven to be 2.25 times faster than $RA^*$ and 17 times faster than the original $A^*$, in average. Moreover, it is more memory-efficient, since it does not need to store a G score matrix.
\end{abstract}




\end{frontmatter}


\section{Introduction}\label{}
The shortest path problem is a well-established and important problem that finds applications in various fields, including robotics (\cite{zhang2018path}, \cite{koubaa2018robot}), VLSI design (\cite{udgirkar2016vlsi}), wireless sensor networks (\cite{radi2012multipath}), and transportation (\cite{bast2016route}, \cite{zhu2015people}). This problem can be represented either on a grid or a more general graph, with different techniques employed for each representation. Grid-based approaches are commonly favored in scenarios like extensive VLSI design, particularly when there are numerous obstacles to navigate around.

Traditionally, two types of methods have been used to address the global path planning problem. The first type is exhaustive search, which involves thorough exploration of the entire search space to find the optimal solution. Examples of exact search algorithms commonly used for this purpose include Dijkstra's algorithm ((\cite{dijkstra1959note}, \cite{dijkstra2022note})) and $A^*$ algorithm (\cite{hart1968formal}). These algorithms guarantee finding the shortest path but may become impractical when dealing with large grids due to their computational complexity.

The second type is local search, which employs metaheuristic algorithms to explore only a portion of the search space. These algorithms, such as Tabu Search (\cite{chaari2014adequacy}), Ant Colony Optimization (\cite{Fatin2020psu}), Genetic Algorithm (\cite{alajlan2013global}), Particle Swarm Optimization (\cite{phung2021safety,huang2023adaptive}), and their multiple variants, provide approximate solutions and are often used when the exhaustive search is not feasible due to the large size of the grid. By exploring a limited subset of the search space, local search algorithms can find reasonably good solutions efficiently.

Recently, some relaxed alternatives of exhaustive search algorithms have emerged. These approaches aim to find a balance between the constraint of achieving optimality in path length and the need for faster search times. The work being discussed here falls into this category. It is demonstrated that the proposed Enhanced Relaxed $A^*$ ($ERA^*$) algorithm is theoretically equivalent to the relaxed $A^*$ ($RA^*$, \cite{ammar2016relaxed}) algorithm in terms of the path length of the solutions it provides. However, it achieves significant savings in terms of time and memory requirements by employing a completely different computation strategy. 

By relaxing the constraint of path length optimality to some extent, this approach manages to expedite the search process while still delivering solutions that are comparable in terms of path length to those obtained through the $RA^*$ algorithm. The key advantage lies in the adoption of a novel computation strategy based on lookup matrices. These matrices store precomputed values that can be efficiently accessed during the search, eliminating the need for redundant calculations and resulting in substantial time and memory savings.

The remaining of this paper is organized as follows: Section \ref{sec:related_works} discussed the most relevant related works. Section \ref{sec:method} presents the proposed $ERA^*$ algorithm and proves its equivalence to $RA^*$ in terms of path cost. Section \ref{sec:eval} presents the results of the evaluation of $ERA^*$ on a benchmark of representative grid maps, and compares it to $A^*$ and $RA^*$ in terms of path cost and execution time. Finally, section \ref{sec:conclusion} concludes the paper and suggests some future works.

\section{Related Works}\label{sec:related_works}

Shortest path algorithms are a crucial tool in the field of computer science and are used to solve a wide range of problems. In general, point-to-point shortest path algorithms work by starting at a source node and exploring the neighboring nodes until they reach the destination node. Along the way, the algorithm keeps track of the shortest distance from the source to each node, using this information to guide the search and ensure that the shortest path is found. There are a variety of shortest path algorithms that have been developed for finding the shortest path between two points in a regular grid (\cite{panda2018survey}). These algorithms typically involve searching through the grid to identify the optimal path, taking into account factors such as the cost of moving from one cell to another and any obstacles that may be present in the grid.

One of the most popular and well-known shortest path algorithms for regular grids is Dijkstra's algorithm, which was first proposed by Dutch computer scientist Edsger Dijkstra (\cite{dijkstra1959note}, \cite{dijkstra2022note}). Dijkstra's algorithm is a general-purpose algorithm that can be used to find the shortest path between any two points in a weighted, directed graph. It is based on the idea of building a "shortest path tree" from the source point to all other points in the grid, with the shortest path to each point being the minimum sum of the weights of the edges along the path. Although Dijkstra's algorithm is relatively simple and easy to implement, it has a time complexity of O($|E|+|V| log |V|$) when using a suitable data structure (Fibonacci heaps), where $|E|$ is the number of edges in the graph and $|V|$ is the number of vertices. This makes it less efficient for larger grids.

Another popular shortest path algorithm for regular grids is $A^*$ \cite{hart1968formal}, which was proposed as an improvement of Dijkstra's algorithm. $A^*$ is a heuristic search algorithm that uses a combination of a best-first search and a cost-estimation function to guide the search towards the goal. Unlike Dijkstra's algorithm, which considers all paths from the source to the goal, $A^*$ only considers paths that are likely to lead to the goal, based on the cost-estimation function. This makes $A^*$ more efficient than Dijkstra's algorithm for many grid-based problems, although the performance of $A^*$ can be sensitive to the choice of cost-estimation function.

\cite{lee1961algorithm} proposed the first Maze-solving algorithm that is based on target-directed grid propagation and is memory-efficient. It was extensively employed in the field of printed circuit board design for finding wire paths. However, \cite{rubin1974lee} later  revealed that the original claim of the algorithm's generality regarding the path cost function is incorrect. 

More recent shortest path algorithms for regular grids include the Jump Point Search (JPS) algorithm, proposed by \cite{harabor2011online} as an improvement over $A^*$ in grid maps. It is based on a selective node expanding process that specifically identifies and expands certain nodes in the grid map, referred to as jump points. Thus, intermediate nodes along a path between two jump points are not expanded at all, which  enhances the speed of $A^*$ search by an order of magnitude.

\cite{hadlock1977shortest} proposed the Minimum Detour (MD) algorithm in G4 regular grids. Its main idea is to calculate the detour number d(P), defined as the number of nodes on the path P that are directed away from the goal. Then, the path length is calculated as M(S,G) + 2d(P), where M(S,G) is the Manhattan distance between the start S and the goal G nodes. This is because any move opposed to the direction of the goal necessarily needs to be compensated to reach the goal. Since M(S,G) is constant for a given (S,G) pair, minimizing the path length is equivalent to minimizing the detour number. Nodes that have lower detour numbers are given higher priority in the grid expansion process, in a breath-first search style. Since then, no similar algorithm has been proposed in G8 grids, because the detour computation is not as simple as in G4 grids. This paper proposes to bridge this gap by introducing a Hadlock-inspired algorithm applicable to G8 grids.

Relaxed Dijkstra and Relaxed $A^*$ ($RA^*$) algorithms proposed by \cite{ammar2016relaxed} exploit the grid-map structure
to establish an accurate approximation of the optimal path, without visiting any cell more than once. The path length is approximated in terms of number of moves on the grid. This approach distinguishes itself from previous bounded relaxation algorithms of $A^*$ (\cite{pohl1970first,pearl1984heuristics,pohl1973avoidance,koll1992new}) primarily by performing relaxation on the exact cost, denoted as g, of the evaluation function f (where f = g + h). This sets it apart from existing relaxations of the $A^*$ algorithm, which typically focus on relaxing the heuristic h. The current work proposes aims to further accelerate the execution of the $RA^*$ algorithm by completely changing its computing paradigm, drawing inspiration from the Hadlock algorithm.

It is worth noting that path planning in a grid map can be substantially accelerated in a completely different way by building connection graphs before applying a search algorithm (\cite{zheng1996finding}). Nevertheless, the fact that the proposed $ERA^*$ algorithm does not construct connection graphs is a significant advantage in many cases. For instance, in large VLSI design problems with a high number of obstacles, the construction of the entire connection graph could be extremely costly.

\section{Methodology}\label{sec:method}

The proposed algorithm is based on calculating detour penalties that are propagated from the source node S to the the goal node G. For a given current node C, we define: 
\begin{itemize}
    \item $\alpha$: the angle between the x-axis and the $\overrightarrow{CG}$ vector.
    \item $\Delta_{x}=x_{G}-x_{C}$, where $x_{G}$ and $x_{C}$ are the abscissas of the goal node and the current node, respectively.
    \item $\Delta_{y}=y_{G}-y_{C}$, where $y_{G}$ and $y_{C}$ are the ordinates of the goal node and the current node, respectively.
    
\end{itemize}

\begin{figure}[!h]  
\begin{center}  
\includegraphics[width=7cm]{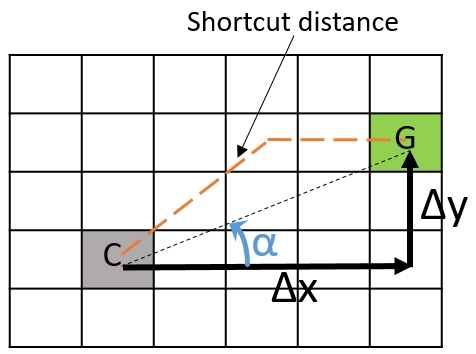}
\caption{\small \sl Example showing the angle $\alpha$ between the current node (C) and the goal node (G), $\Delta_{x}=x_{G}-x_{C}$, $\Delta_{y}=y_{G}-y_{C}$, and the shortcut distance between C and G, which corresponds to the minimum distance in a G8 grid when there are no obstacles between the two nodes.
\label{fig:alpha_angle}}  
\end{center}  
\end{figure}

\begin{figure}[H]  
\begin{center}  
\includegraphics[width=17cm]{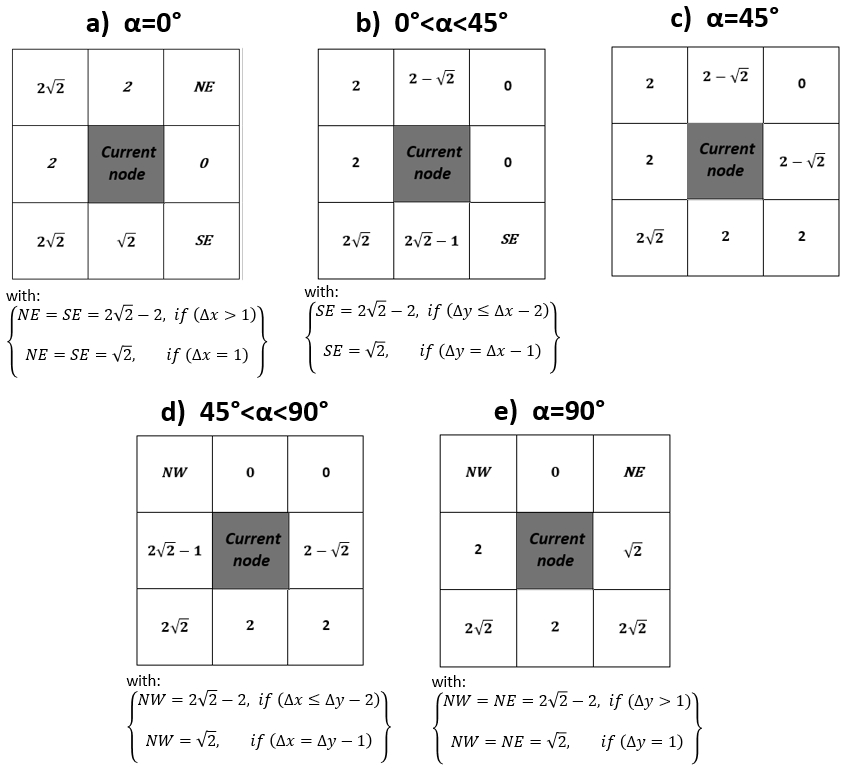}
\caption{\small \sl Incremental detour penalty lookup matrices for angles between the current node and the goal node from 0° to 90°. The remaining cases can be obtained from the above cases, using simple 90° rotations of the 3x3 matrix. NE, SE, and NW stand for Northeast, Southeast, and Northwest, respectively. These incremental penalties are added when propagating the total penalty the source node to the goal node.
\label{fig:cases}}  
\end{center}  
\end{figure}

Figure \ref{fig:alpha_angle} clarifies the definition of the variables $\alpha$, $\Delta_{x}$, and $\Delta_{y}$ on an example. Figure \ref{fig:cases} presents the five first matrices of incremental detour penalty for $\alpha \in [0^{\circ},90^{\circ}]$. These 3x3 matrices store the incremental detour (penalty) from the goal  expressed as: 
\begin{equation}\label{eq_D_definition}
D(n_{i}) - D(n_{i-1}) = dist(n_{i-1}, n_{i})+h(n_{i})-h(n_{i-1})
\end{equation}

Where $h$ is the shortcut distance to goal (shortest path assuming there are no obstacles, see figure \ref{fig:alpha_angle}).

The value of some penalties, which are denoted as NE (North-East), SE (South-East), and NW (North-West), depend on the value of $\Delta_{x}$ and $\Delta_{y}$. For instance, in case (a), corresponding to $\alpha=0$ (C and G on the same row, with G on the right, so that the optimal path between them is equal to $\Delta_{x}$), if $\Delta_{x}=1$, it means that the goal is adjacent to the current node on its right. Consequently, if we move towards NE or SE, we will reach the goal with a minimum cost of $\sqrt{2}+1$ instead of just $1$ for the optimal path between C and G. That is why the penalty is $\sqrt{2}$. Whereas, for all other values of $\Delta_{x}>1$, if we move towards NE, we can return back to the row containing G with a total cost of $2\sqrt{2}$ instead of $2$ for the optimal path. That is why the penalty is $2\sqrt{2}-2$. Obviously, the case $\Delta_{x}<1$ cannot correspond to angle $\alpha=0$. The reasoning is similar for the other cases.

Notice that the matrix (e) can be obtained from the matrix (a) by applying a 90° anti-clockwise rotation and substituting $\Delta_{x}$ for $\Delta_{y}$ in the formulas of NE/SE that becomes NW/NE. Likewise:

\begin{itemize}
    \item (f) $90\degree<\alpha<135\degree$: is obtained from (b) by applying a 90° anti-clockwise rotation.
    \item (g) $\alpha=135\degree$: is obtained from (c) by applying a 90° anti-clockwise rotation.
    \item (h)	$135\degree<\alpha<180\degree$: is obtained from (d) by applying a 90° anti-clockwise rotation.
\end{itemize}

In total, we have 28 fixed $3\times3$ matrices (counting the cases of different $\Delta_{x}$ and $\Delta_{y}$ values), that we store before running the search algorithm.

As in \cite{hadlock1977shortest}, nodes that have lower detour numbers are given higher priority in the grid expansion process. This process proceeds until the goal position is reached. Then the path is reconstructed backwards from the goal cell, by moving from each cell to its parent cell that expanded it, until the starting cell is reached.


Instead of two matrices (actual cost $g$ and combined score $f$) as in $A^*$ and $RA^*$, only one matrix of detour penalties ($D$) is stored in memory. At a give node $n_{i}$, the detour value can be calculated recursively, using the euclidean distance $dist$ and the heuristic function $h$ (shortcut distance), as:

\begin{equation} 
\begin{split}
D(n_{i}) & = D(n_{i-1})+dist(n_{i-1}, n_{i})+h(n_{i})-h(n_{i-1}) \\
 & = D(n_{i-2})+dist(n_{i-2}, n_{i-1})+dist(n_{i-1}, n_{i})+h(n_{i-1})-h(n_{i-2})+h(n_{i})-h(n_{i-1}) \\
& = ... \\
& = D(n_{0})+\sum_{k=0}^{i-1} {dist(n_{k},n_{k+1})}+\sum_{k=1}^{i} {h(n_{k})-h(n_{k-1})}\nonumber
\end{split}
\end{equation}

With:
\begin{equation}
\begin{split}
& D(n_{0})=0\\
& \sum_{k=0}^{i-1} {dist(n_{k},n_{k+1})}=g(n_{i})\\
& \sum_{k=1}^{i} {h(n_{k})-h(n_{k-1})}=h(n_{i})-h(n_{0})\nonumber
\end{split}
\end{equation}

Therefore:

\begin{equation} 
\begin{split}
D(n_{i}) & = g(n_{i})+h(n_{i})-h(n_{0}) \\
& = f(n_{i})-h(n_{0})
\end{split}
\end{equation}

Consequently, since $h(n_{0})$ is constant, optimizing $D$ is equivalent to optimizing $f$.

\begin{algorithm}[!h]
\label{search_algo}
\SetKwInOut{Input}{input}\SetKwInOut{Output}{output}
\Input{$map$: The whole grid with obstacle information\\
$max\_nb\_iter$: maximum number of iterations\\
$(i_{S},j_{S})$: coordinates of the Start node S on the grid\\
$(i_{G},j_{G})$: coordinates of the Goal node G on the grid}
\Output{$D$: penalty matrix\\
$nb\_iter$: number of iterations to find the near-optimal path}
\BlankLine
\SetAlgoLined
$nb\_iter$ = 0\\
$fail$ = False \\
Store the 28 $(3\times3)$ $D_{i}$ penalty matrices \quad\textit{ //  See Fig.\ref{fig:cases} and eq.\ref{eq_D_definition}}\\
$I = [(i_{S},j_{S},0)]$  \quad \quad \textit{// Priority queue of expanded nodes} \\
\textit{ // Initialization of the distance matrix D and the predecessors matrix P on the map:} \\
\ForEach{node in $map$}  
{$D(node) =  +\infty$ \\
$P(node) = +\infty$} 
$D(i_{S},j_{S}) = 0$ \\
\While{($D(i_{G},j_{G}) = +\infty$ \textbf{and} $nb\_iter<max\_nb\_iter$ \textbf{and} $size(I)>0$)}
{
$(i_{C},j_{C})$ = dequeueMin(I) \quad \textit{// C will be the current node}\\
$\Delta x = i_{G} - i_{C}$\\
$\Delta y = j_{G} - j_{C}$\\
Choose the correct $(3\times3)$ $D_{i}$ penalty matrix $matP$ according to the relative value of $\Delta x$ and $\Delta y$\\
$J$ = \{$(i,j,D(i,j))$ $\vert$ $(i,j)=N\in map$,  N neighbor of C, N is not an obstacle, and $D(i,j)=+\infty$\}\\
\ForEach{$(i_{P},j_{P},D(i_{P},j_{P}))$ in $J$}  
{$P(i_{P},j_{P})=(i_{C},j_{C})$\\
$D(i_{P},j_{P})=D(i_{C},j_{C})+matP(i_{P}-i_{C},j_{P}-j_{C})$\\
}
Enqueue J in I\\
$nb\_iter$++
}
 \caption{$ERA^*$ search algorithm}
\end{algorithm}

Algorithm \ref{search_algo} presents the main steps for path exploration in $ERA^*$. The algorithm takes as input a grid represented by a 2D matrix ($map$) where obstacles are marked with a predefined value, and the coordinates of the start (S) and goal (G) nodes. It calculates the penalty matrix D with a finite value for all explored nodes. The algorithm stops when the node exploration reaches G, or when a predefined maximum number of iterations is reached. We first store the 28 $3\times3$ incremental penalty matrices (see Figure \ref{fig:cases}), and initialize the priority queue I of expanded nodes to the start node S with associated penalty 0 (line 4). The penalty matrix D and the Predecessor matrix P (used for path reconstruction) are both initialized with default infinity values (lines 6-8), except for the penalty associated with S which is initialized to 0 (line 10). Then, we iterate over the priority queue I by dequeuing each time the node that has the minimum penalty value $D(i,j)$. Based on the value of $\Delta x$ and $\Delta y$ between this current node C and the goal node, we choose the corresponding incremental penalty matrix $D_{i}$. This choice is accomplished through a series of if statements where the most likely cases (e.g., inequalities such as $\Delta y \ge 1$) are placed before less likely cases (e.g., equalities such as $\Delta x = 0$), in order to optimize execution time. We define J (line 16) as the set of all free (i.e., non-obstacle) neighbor nodes of C for which the penalty value $D(i,j)$ is infinite (i.e., has not been calculated yet). For each node in J, we save its predecessor C in the P matrix (line 17) for later path reconstruction, and we calculate its penalty value $D(i_{P},j_{P})$ as the sum of the current node's penalty plus the corresponding incremental penalty in the $3\times3$ $D_{i}$ matrix (line 18). Then, we enqueue J in I (line 21), where the priorities correspond to the D values. If there is a path between S and G, G will be reached and its penalty value calculated, unless the predefined maximum number of iteration is attained.

Once Algorithm \ref{search_algo} reaches the goal G, Algorithm \ref{reconstruction_algo} is used to reconstruct the path backward. It takes as input the penalty matrix D and predecessors matrix P that were both generated in Algorithm \ref{search_algo}, and the start and goal coordinates on the grid. If $D(i_{G},j_{G}) \neq +\infty$ (line 1), it means that Algorithm \ref{search_algo} has reached the goal and a path from S to G has been found. The path reconstruction algorithm starts from the goal $(i_{G},j_{G})$ (line 3), and prepends the current node's predecessor to the path at each iteration (line 14), while adding $1$ (for horizontal and vertical moves) or $\sqrt{2}$ (for diagonal moves) until it reaches the start node S. Alternatively, we can count the number of diagonal moves, and multiply it by $\sqrt{2}$ only at the end to avoid accumulating numerical inaccuracies.

\begin{algorithm}[!h]
\label{reconstruction_algo}
\SetKwInOut{Input}{input}\SetKwInOut{Output}{output}
\Input{$D$: penalty matrix\textit{  \quad\quad\quad  // Output from Algorithm 1}\\
$P$: Predecessors matrix\textit{  \quad  // Output from Algorithm 1}\\
$(i_{S},j_{S})$: coordinates of the Start node S on the grid\\
$(i_{G},j_{G})$: coordinates of the Goal node G on the grid}
\Output{$path$: near-optimal path between S and G\\
$length$: length of the near-optimal path between S and G\\
$fail:$ True if no path has been found}
\BlankLine
\SetAlgoLined
\uIf{$D(i_{G},j_{G}) \neq +\infty$}
{
$fail$ = False\\
$(i_{C},j_{C})$ = $(i_{G},j_{G})$\textit{   \quad // Path reconstruction starts from the Goal}\\
$path$ = $[(i_{G},j_{G})]$\\
$length=0$\\
\While{$(i_{C},j_{C}) \neq (i_{S},j_{S})$}
{
$(i_{P},j_{P})$ = $P(i_{C},j_{C})$\\
\uIf{$i_{P}=i_{C}$ \textbf{or} $j_{P}=j_{C}$}{
    $length$ += 1
  }
  \Else{
    $length$ += $\sqrt{2}$
  }
  $(i_{C},j_{C})$ = $(i_{P},j_{P})$\\
  $path.prepend((i_{C},j_{C}))$
}}
\Else{
$fail$ = True
}

 \caption{path reconstruction algorithm}
\end{algorithm}


\section{Evaluation and discussion}\label{sec:eval}

\subsection{Dataset}
The simulation study was performed using Matlab R2012a on a laptop with an Intel Core i7 processor (2.4 GHz) and 16 GB of RAM. The evaluation of the algorithms was based on the same benchmark used in \cite{ammar2016relaxed}, which is composed of four categories of maps: 
\begin{itemize}
    \item 26 maps with randomly placed rectangular obstacles of various obstacle sizes and ratios, ranging from 100 x 100 to 2000 x 2000 in size.
    \item  6 mazes with passages of different sizes, all 512 x 512 in size, with variable corridor size.
    \item  4 room maps (512 x 512) filled with random square rooms of variable size. 
    \item  6 maps from video games and 1 real-world map (Willow Garage), ranging from 512 x 512 to 1024 x 1024 in size and selected for their varying levels of difficulty.
\end{itemize}

We designed and generated the first category of random maps, while the other three categories (mazes, rooms, and video games) were selected from a large set of benchmarking maps provided by \cite{sturtevant2012benchmarks}. For each map, we conducted 30 runs with randomly selected start and goal nodes each time. Thus, the total number of runs is 1290 (43 maps × 30) for each algorithm.

We compared the proposed $ERA^*$ algorithm to the original A$_{t}^{*}$ ($A^*$ using path tie-breaks) and to $RA^*$$_{wot}$ (Relaxed $A^*$ without using path tie-breaks) which was proven in \cite{ammar2016relaxed} to provide the best tradeoff between path cost and execution time among the tested relaxed variants in G8 grids.

\subsection{Results}
Two main performance metrics are considered to evaluate the three global planners:

\begin{itemize}
\item The path length: it represents the length of the shortest global path found by the planner.
\item The execution time: time spent by an algorithm to find its best (or optimal) solution.
\end{itemize}


\begin{table}[h]
\centering
\resizebox{\textwidth}{!}{\begin{tabular}{|l|l|l|l|l|l|l|l|l|}
\hline
Algorithm       & 100x100  & 500x500  & 1000x1000 & 2000x2000 & \begin{tabular}[c]{@{}l@{}}Mazes\\(512x512)\end{tabular} & \begin{tabular}[c]{@{}l@{}}Rooms\\(512x512)\end{tabular} & \begin{tabular}[c]{@{}l@{}}VideoGames\\(512x512 to \\1024x1024)\end{tabular} & All       \\ 
\hline
$A^*_{t}$    & 100.0\% & 100.0\% & 100.0\%  & 100.0\%  & 100.0\%                                                  & 100.0\%                                                  & 100.0\%                                                                      & 100.0\%  \\ 
\hline
$RA^*_{wot}$ & 79.6\%  & 85.0\%  & 78.9\%   & 96.7\%   & 21.7\%                                                   & 25.0\%                                                   & 47.2\%                                                                       & 62.9\%   \\ 
\hline
$ERA^*$            & 81.9\%  & 86.0\%  & 80.6\%   & 96.7\%   & 4.4\%                                                    & 23.3\%                                                   & 46.7\%                                                                       & 61.1\%   \\
\hline
\end{tabular}}
\caption{Percentage of exact optimal paths (when a path exists) per environment, for the three tested algorithms.}
\label{tab:percent_optimal_paths} 
\end{table}

Table \ref{tab:percent_optimal_paths} shows the percentage of optimal paths (when a path exists) for the three tested algorithms, per environment. The original $A^*$ provides the optimal in all cases, since no relaxation is applied to it. We observe that $ERA^*$ yields a slightly higher rate of optimal paths than $RA^*_{wot}$ for random environments of different sizes, a slightly lower rate for rooms and video games, and a markedly lower rate for mazes, in which path lengths are the longest in average, as shown in table \ref{tab:avg_path_cost}. This is due to the numerical issues entailed by the incremental computation. Nevertheless, table \ref{tab:percent_extra_length} shows that the percentage of extra length compared to the optimal path remains low at 2.3\% in average, and a maximum of 10.4\%, which is slightly lower than the maximum extra length produced by $RA^*_{wot}$.

\begin{table}[!h]
\centering
\resizebox{\textwidth}{!}{\begin{tabular}{|l|l|l|l|l|l|l|l|l|} 
\hline
Algorithm       & 100x100 & 500x500 & 1000x1000 & 2000x2000 & \begin{tabular}[c]{@{}l@{}}Mazes\\(512x512)\end{tabular} & \begin{tabular}[c]{@{}l@{}}Rooms\\(512x512)\end{tabular} & \begin{tabular}[c]{@{}l@{}}VideoGames\\(512x512 to\\1024x1024)\end{tabular} & All    \\ 
\hline
$A^*_{t}$    & 60.4    & 284.8   & 631.2     & 1086.2    & 1479.1                                                   & 317.1                                                    & 375.8                                                                       & 490.3  \\ 
\hline
$RA^*_{wot}$ & 60.7    & 285     & 632.3     & 1086.3    & 1501                                                     & 321                                                      & 381.3                                                                       & 494.8  \\ 
\hline
$ERA^*$            & 60.7    & 285.2   & 633       & 1086.3    & 1517                                                     & 323.6                                                    & 382.8                                                                       & 497.7  \\
\hline
\end{tabular}}
\caption{Average path cost per environment size, in cell units.}
\label{tab:avg_path_cost} 
\end{table}

\begin{table}[!h]
\centering
\begin{tabular}{|l|l|l|l|} 
\hline
Algorithm & Mean   & Std    & Max      \\ 
\hline
$RA^*_{wot}$    & 1.7\% & 1.6\% & 10.7\%  \\ 
\hline
$ERA^*$      & 2.3\% & 1.8\% & 10.4\%  \\
\hline
\end{tabular}
\caption{Percentage of extra length compared to optimal path, calculated for non-optimal paths over all environments.}
\label{tab:percent_extra_length} 
\end{table}

\begin{figure}[!h]  
\begin{center}  
\hspace*{-1in}
\includegraphics[width=14cm]{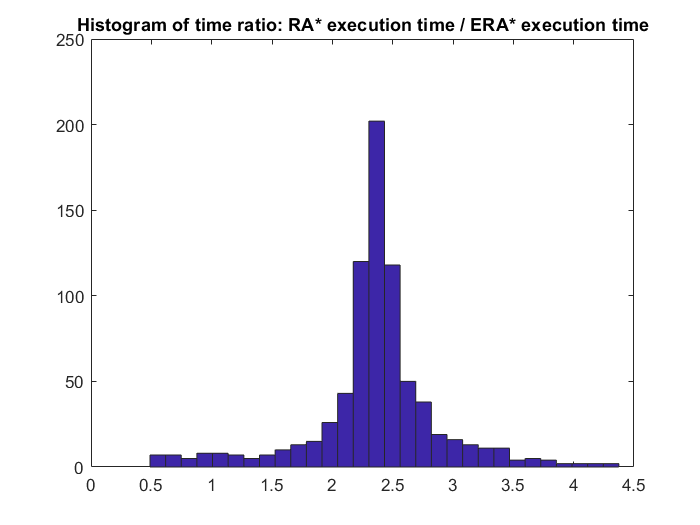}
\caption{\small \sl Histogram of the execution time ratio between $RA^*_{wot}$ and $ERA^*$ algorithms.
\label{fig:hist_exec_time_ratio}}  
\end{center}  
\end{figure}

On the other hand, Table \ref{tab:avg_exec_time} shows the average execution time of each algorithm in each grid environment. $ERA^*$ is 2.25 times faster in average than $RA^*_{wot}$. This ratio varies from 1.74$\times$ to 2.47$\times$ depending on the type of grid environment. Figure \ref{fig:hist_exec_time_ratio} displays the histogram of this ratio. It shows that $RA^*_{wot}$ is faster than $ERA^*$ in only very few cases. 

\begin{table}[h]
\centering
\resizebox{\textwidth}{!}{\begin{tabular}{|l|l|l|l|l|l|l|l|l|} 
\hline
Algorithm       & 100x100 & 500x500 & 1000x1000 & 2000x2000 & \begin{tabular}[c]{@{}l@{}}Mazes\\(512x512)\end{tabular} & \begin{tabular}[c]{@{}l@{}}Rooms\\(512x512)\end{tabular} & \begin{tabular}[c]{@{}l@{}}VideoGames\\(512x512 to\\1024x1024)\end{tabular} & All    \\ 
\hline
$A^*_{t}$    & 0.23    & 3.71    & 39.27     & 113.96    & 113                                                      & 17.76                                                    & 29.97                                                                       & 31.35  \\ 
\hline
$RA^*_{wot}$ & 0.06    & 1.15    & 6.33      & 24.82     & 11.04                                                    & 2.87                                                     & 3.5                                                                         & 4.12   \\ 
\hline
$ERA^*$            & 0.03    & 0.51    & 2.71      & 10.05     & 4.65                                                     & 1.3                                                      & 2.01                                                                        & 1.83   \\
\hline
\end{tabular}}
\caption{Average execution time (in seconds).}
\label{tab:avg_exec_time} 
\end{table}

\begin{figure}[!h]  
\begin{center}  
\hspace*{-2in}
\includegraphics[width=24cm]{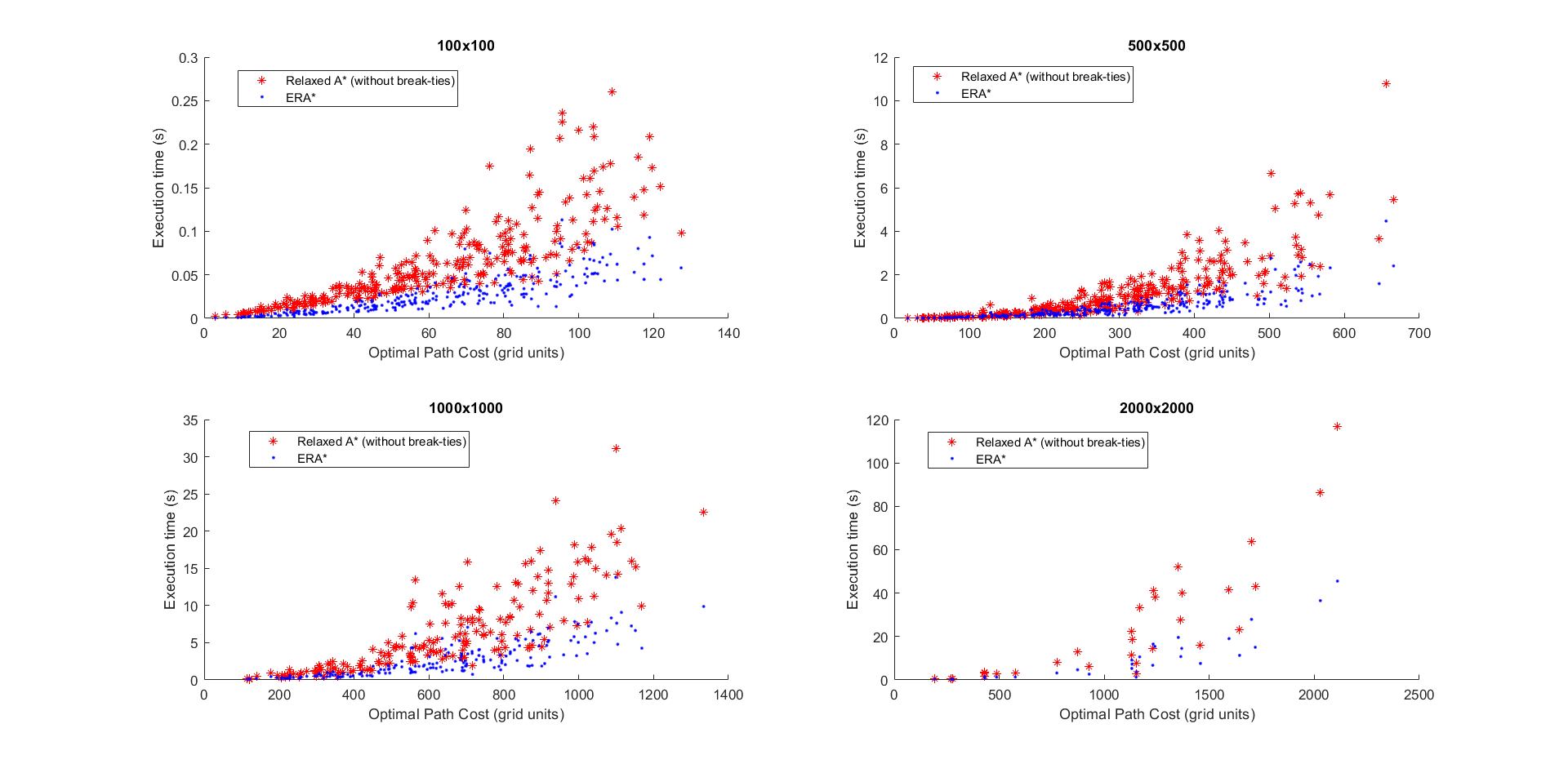}
\caption{\small \sl Comparison of execution time between the proposed $ERA^*$ (in blue) and $RA^*_{wot}$ (in red) algorithms, for different environment sizes (from 100x100 to 2000x2000 nodes), randomly generated.
\label{fig:exec_random_env}}  
\end{center}  
\end{figure}

\begin{figure}[!h]  
\begin{center}  
\hspace*{-2in}
\includegraphics[width=22cm]{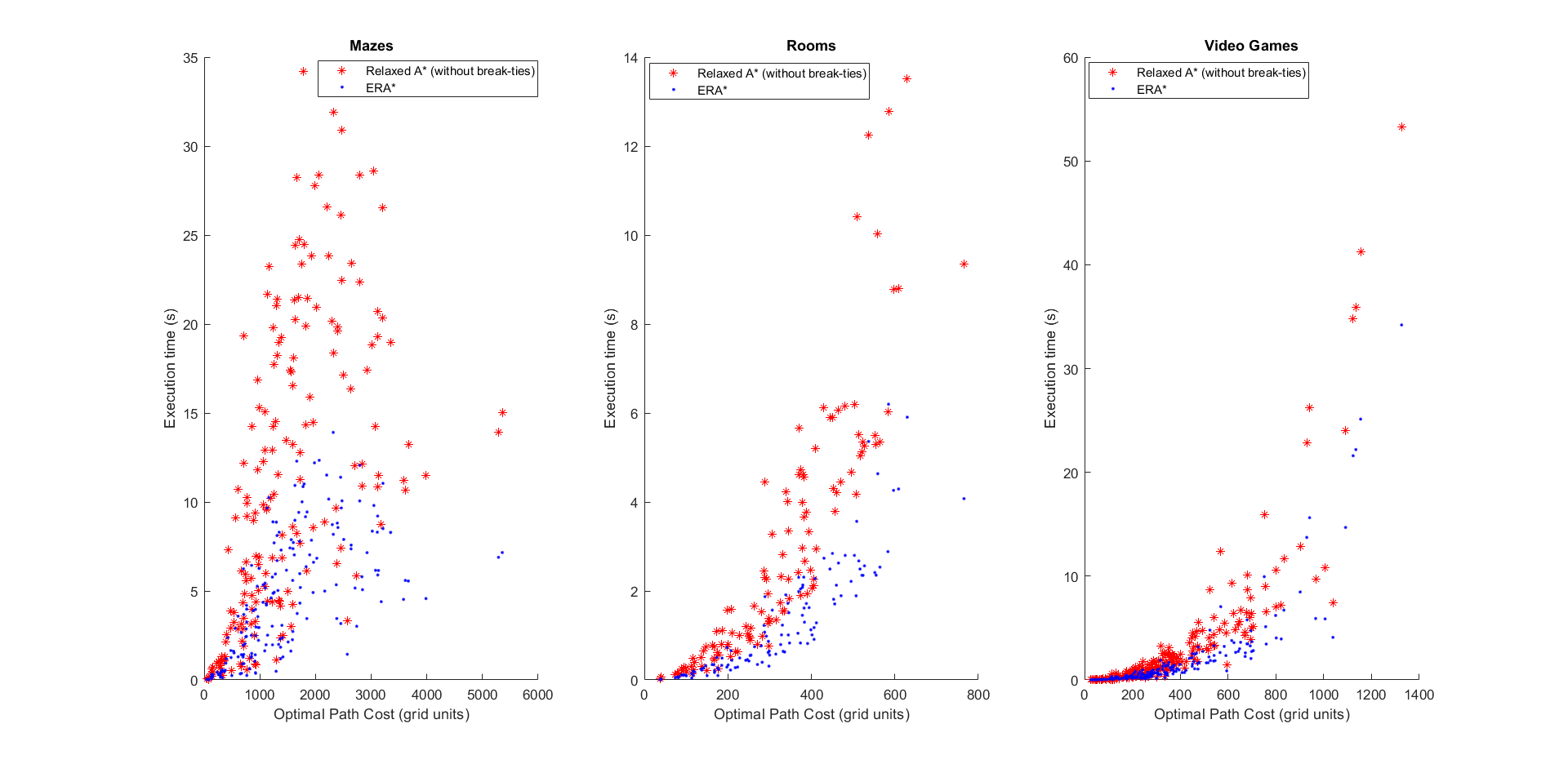}
\caption{\small \sl Comparison of execution time between the proposed $ERA^*$ (in blue) and $RA^*_{wot}$ (in red) algorithms, for different structured environments.
\label{fig:exec_struct_env}}  
\end{center}  
\end{figure}

When compared to $A^*$,  $ERA^*$ is 17.13 times faster in average, with a range from 7.27$\times$ to 24.3$\times$. Table \ref{tab:percent_rank} shows that $ERA^*$ has consistently the fastest execution time among the three algorithms in nearly 90\% of cases. This superiority applies in all tested grid environments as can be seen in the scatter plots in Figure \ref{fig:exec_random_env} (randomly generated environments of various sizes) and Figure \ref{fig:exec_struct_env} (structured environments) which compare $ERA^*$ and $RA^*_{wot}$ in the Cost/Time space. These two figures also show that the range of path length for the two algorithms is similar.

\begin{table}[h]
\centering
\caption{Percentage of runs for which each algorithm appears in rank 1 to 3, with regards to the execution time.}
\begin{tabular}{|l|c|c|c|} 
\hline
Rank & 1 & 2 & 3\\
\hline
$A^*_{t}$    & 9.0\%   & 5.3\%                         & 85.7\%   \\ 
\hline
$RA^*_{wot}$ & 1.1\%   & 86.6\%                        & 12.3\%   \\ 
\hline
$ERA^*$            & 89.9\%  & 8.1\%                         & 2.0\%    \\
\hline
\end{tabular}
\label{tab:percent_rank} 
\end{table}

\begin{figure}[h]  
\begin{center}  
\includegraphics[width=9cm]{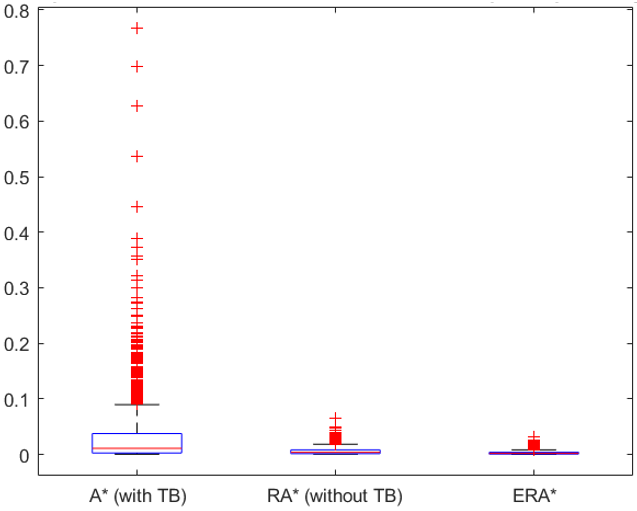}
\caption{\small \sl Box plot of the  execution time divided by the optimal path length, for $A^*$, $RA^*_{wot}$ and $ERA^*$ algorithms.
\label{fig:box_plot_time_ratio}}  
\end{center}  
\end{figure}

\begin{figure}[!h]  
\begin{center}  
\includegraphics[width=8cm]{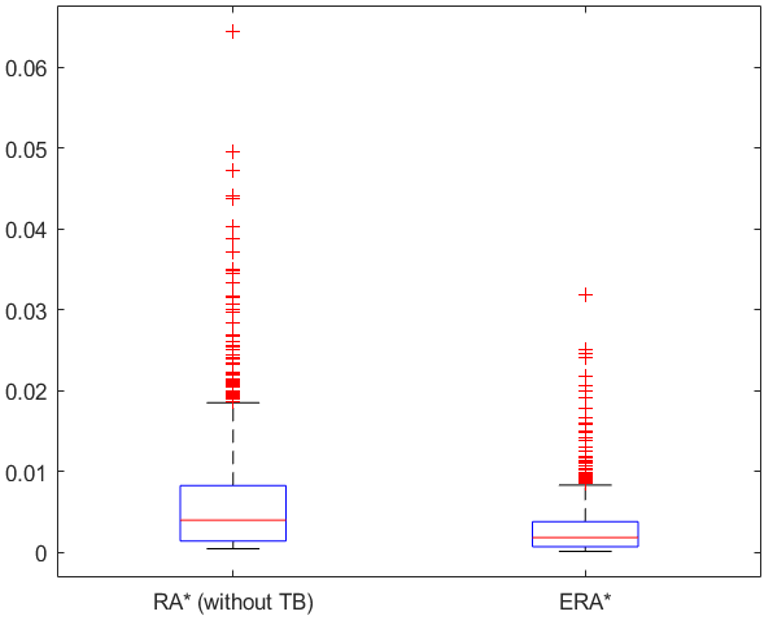}
\caption{\small \sl Box plot of the   execution time divided by the optimal path length, for $RA^*_{wot}$ and $ERA^*$ algorithms.
\label{fig:box_plot_time_ratio_2algos}}  
\end{center}  
\end{figure}

Figure \ref{fig:box_plot_time_ratio} depicts the box plot of the ratio between the execution time and the optimal path for the three algorithms. It shows a markedly reduced range for $ERA^*$ and  $RA^*$ compared to $A^*$. On a closer scale, Figure \ref{fig:box_plot_time_ratio_2algos} shows that $ERA^*$ further reduces this range by almost one half compared to $RA^*$. 

\begin{figure}[!h]  
\begin{center}  
\includegraphics[width=9cm]{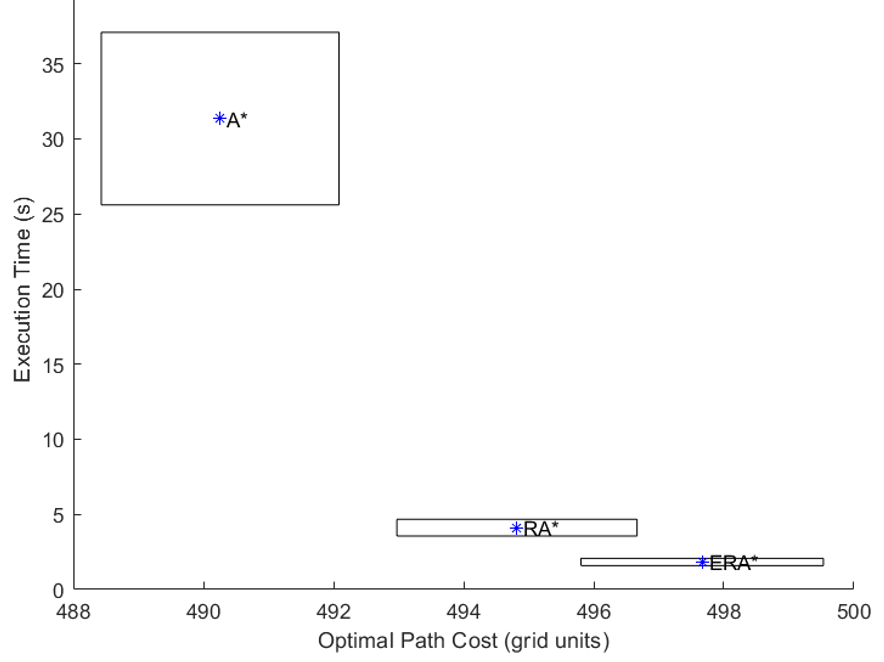}
\caption{\small \sl Comparison between the tested algorithms in cost/time space, in terms of average and standard deviation. The width of each rectangle is proportional to the path cost standard deviation, and its height is proportional to the execution time standard deviation. The average is represented by a star at the center of each rectangle.
\label{fig:compare_2d}}  
\end{center}  
\end{figure}

Figure \ref{fig:compare_2d} provides a comprehensive evaluation of the performance of the three algorithms by representing them in cost/time space, in terms of average and standard deviation. The aim of this comparison is to evaluate the performance of different algorithms for both the path cost and the execution time. The cost/time space, therefore, shows the trade-off between the cost and time incurred by each algorithm. The evaluation is done based on two metrics: average and standard deviation. The width of each rectangle is proportional to the path cost standard deviation, while the height is proportional to the execution time standard deviation. The average is represented by a star at the center of each rectangle. The figure shows that $ERA^*$ provides the best tradeoff between cost and time, a reduced range of execution time, and a range of path cost that is close to $RA^*$'s.

Furthermore, we conducted an analysis on the cost and time performance of the three algorithms on the described set of experiments, using a T-test. The first comparison was between the $ERA^*$ and $RA^*_{wot}$ algorithms, where the null hypothesis of equal means for time was rejected with a very low p-value of $1.1216e^{-20}$. Whereas, the null hypothesis for cost was accepted with a p-value of 0.91, indicating that the difference in cost between the two algorithms is not statistically significant. The second comparison was between the $ERA^*$ and $A^*_{t}$ algorithms. Again, the null hypothesis for equal means in time was rejected with a very low p-value of $1.1002e^{-35}$, indicating that there is a statistically significant difference in time performance between the two algorithms. While the null hypothesis for cost was accepted with a p-value of 0.7607, indicating that the cost difference between the two algorithms is not statistically significant. 



\section{Conclusion}\label{sec:conclusion}
In conclusion, this paper introduced a novel algorithm for solving the point-to-point shortest path problem on a static regular 8-neighbor connectivity (G8) grid. The algorithm is a generalized version of the Hadlock algorithm specifically designed for G8 grids. 
This work falls within the category of relaxed alternative algorithms that balance the trade-off between path length optimality and search time. By relaxing the constraint of path length optimality to some extent, the proposed Enhanced Relaxed A* (ERA*) algorithm accelerates the search process while still providing solutions with similar path lengths to those obtained using RA*. The key advantage lies in the novel computation strategy utilizing lookup matrices, which significantly reduces redundant calculations, resulting in substantial time and memory savings.

Experimental results obtained through extensive testing on various grid maps validate the algorithm's effectiveness. On average, it is 2.25 times faster than RA* and 17 times faster than the original A* algorithm. Additionally, it exhibits improved memory efficiency since it eliminates the need for storing a G score matrix.

Future research can focus on exploring further optimizations and extensions of the ERA* algorithm. This could involve investigating its performance in dynamic or uncertain environments, incorporating additional constraints or cost factors, and exploring possibilities for parallelization or distributed computation. Additionally, the algorithm's applicability to other grid types or graph structures can be explored. The insights gained from this work lay the foundation for potential advancements in efficient path planning algorithms for various domains and applications.










\clearpage
\bibliographystyle{./elsarticle-num-names.bst}
\bibliography{biblio}

\begin{thebibliography}{26}
\expandafter\ifx\csname natexlab\endcsname\relax\def\natexlab#1{#1}\fi
\providecommand{\url}[1]{\texttt{#1}}
\providecommand{\href}[2]{#2}
\providecommand{\path}[1]{#1}
\providecommand{\DOIprefix}{doi:}
\providecommand{\ArXivprefix}{arXiv:}
\providecommand{\URLprefix}{URL: }
\providecommand{\Pubmedprefix}{pmid:}
\providecommand{\doi}[1]{\href{http://dx.doi.org/#1}{\path{#1}}}
\providecommand{\Pubmed}[1]{\href{pmid:#1}{\path{#1}}}
\providecommand{\bibinfo}[2]{#2}
\ifx\xfnm\relax \def\xfnm[#1]{\unskip,\space#1}\fi
\bibitem[{Zhang et~al.(2018)Zhang, Lin, and Chen}]{zhang2018path}
\bibinfo{author}{H.-y. Zhang}, \bibinfo{author}{W.-m. Lin},
  \bibinfo{author}{A.-x. Chen},
\newblock \bibinfo{title}{Path planning for the mobile robot: A review},
\newblock \bibinfo{journal}{Symmetry} \bibinfo{volume}{10}
  (\bibinfo{year}{2018}) \bibinfo{pages}{450}.
\bibitem[{Koub{\^a}a et~al.(2018)Koub{\^a}a, Bennaceur, Chaari, Trigui, Ammar,
  Sriti, Alajlan, Cheikhrouhou, and Javed}]{koubaa2018robot}
\bibinfo{author}{A.~Koub{\^a}a}, \bibinfo{author}{H.~Bennaceur},
  \bibinfo{author}{I.~Chaari}, \bibinfo{author}{S.~Trigui},
  \bibinfo{author}{A.~Ammar}, \bibinfo{author}{M.-F. Sriti},
  \bibinfo{author}{M.~Alajlan}, \bibinfo{author}{O.~Cheikhrouhou},
  \bibinfo{author}{Y.~Javed}, \bibinfo{title}{Robot path planning and
  cooperation}, volume \bibinfo{volume}{772}, \bibinfo{publisher}{Springer},
  \bibinfo{year}{2018}.
\bibitem[{Udgirkar and Indumathi(2016)}]{udgirkar2016vlsi}
\bibinfo{author}{G.~Udgirkar}, \bibinfo{author}{G.~Indumathi},
\newblock \bibinfo{title}{Vlsi global routing algorithms: A survey},
\newblock in: \bibinfo{booktitle}{2016 3rd International Conference on
  Computing for Sustainable Global Development (INDIACom)},
  \bibinfo{organization}{IEEE}, \bibinfo{year}{2016}, pp.
  \bibinfo{pages}{2528--2533}.
\bibitem[{Radi et~al.(2012)Radi, Dezfouli, Bakar, and Lee}]{radi2012multipath}
\bibinfo{author}{M.~Radi}, \bibinfo{author}{B.~Dezfouli},
  \bibinfo{author}{K.~A. Bakar}, \bibinfo{author}{M.~Lee},
\newblock \bibinfo{title}{Multipath routing in wireless sensor networks: survey
  and research challenges},
\newblock \bibinfo{journal}{sensors} \bibinfo{volume}{12}
  (\bibinfo{year}{2012}) \bibinfo{pages}{650--685}.
\bibitem[{Bast et~al.(2016)Bast, Delling, Goldberg, M{\"u}ller-Hannemann,
  Pajor, Sanders, Wagner, and Werneck}]{bast2016route}
\bibinfo{author}{H.~Bast}, \bibinfo{author}{D.~Delling},
  \bibinfo{author}{A.~Goldberg}, \bibinfo{author}{M.~M{\"u}ller-Hannemann},
  \bibinfo{author}{T.~Pajor}, \bibinfo{author}{P.~Sanders},
  \bibinfo{author}{D.~Wagner}, \bibinfo{author}{R.~F. Werneck},
\newblock \bibinfo{title}{Route planning in transportation networks},
\newblock \bibinfo{journal}{Algorithm engineering: Selected results and
  surveys}  (\bibinfo{year}{2016}) \bibinfo{pages}{19--80}.
\bibitem[{Zhu and Levinson(2015)}]{zhu2015people}
\bibinfo{author}{S.~Zhu}, \bibinfo{author}{D.~Levinson},
\newblock \bibinfo{title}{Do people use the shortest path? an empirical test of
  wardrop’s first principle},
\newblock \bibinfo{journal}{PloS one} \bibinfo{volume}{10}
  (\bibinfo{year}{2015}) \bibinfo{pages}{e0134322}.
\bibitem[{Dijkstra(1959)}]{dijkstra1959note}
\bibinfo{author}{E.~W. Dijkstra},
\newblock \bibinfo{title}{A note on two problems in connexion with
  graphs:(numerische mathematik, 1 (1959), p 269-271)}  (\bibinfo{year}{1959}).
\bibitem[{Dijkstra(2022)}]{dijkstra2022note}
\bibinfo{author}{E.~W. Dijkstra},
\newblock \bibinfo{title}{A note on two problems in connexion with graphs},
\newblock in: \bibinfo{booktitle}{Edsger Wybe Dijkstra: His Life, Work, and
  Legacy}, \bibinfo{year}{2022}, pp. \bibinfo{pages}{287--290}.
\bibitem[{Hart et~al.(1968)Hart, Nilsson, and Raphael}]{hart1968formal}
\bibinfo{author}{P.~E. Hart}, \bibinfo{author}{N.~J. Nilsson},
  \bibinfo{author}{B.~Raphael},
\newblock \bibinfo{title}{A formal basis for the heuristic determination of
  minimum cost paths},
\newblock \bibinfo{journal}{IEEE transactions on Systems Science and
  Cybernetics} \bibinfo{volume}{4} (\bibinfo{year}{1968})
  \bibinfo{pages}{100--107}.
\bibitem[{Ch{\^a}ari et~al.(2014)Ch{\^a}ari, Koub{\^a}a, Bennaceur, Ammar,
  Trigui, Tounsi, Shakshuki, and Youssef}]{chaari2014adequacy}
\bibinfo{author}{I.~Ch{\^a}ari}, \bibinfo{author}{A.~Koub{\^a}a},
  \bibinfo{author}{H.~Bennaceur}, \bibinfo{author}{A.~Ammar},
  \bibinfo{author}{S.~Trigui}, \bibinfo{author}{M.~Tounsi},
  \bibinfo{author}{E.~Shakshuki}, \bibinfo{author}{H.~Youssef},
\newblock \bibinfo{title}{On the adequacy of tabu search for global robot path
  planning problem in grid environments},
\newblock \bibinfo{journal}{Procedia Computer Science} \bibinfo{volume}{32}
  (\bibinfo{year}{2014}) \bibinfo{pages}{604--613}.
\bibitem[{Ajeil et~al.(2020)Ajeil, Ibraheem, Azar, and Humaidi}]{Fatin2020psu}
\bibinfo{author}{F.~H. Ajeil}, \bibinfo{author}{I.~K. Ibraheem},
  \bibinfo{author}{A.~T. Azar}, \bibinfo{author}{A.~J. Humaidi},
\newblock \bibinfo{title}{Grid-based mobile robot path planning using
  aging-based ant colony optimization algorithm in static and dynamic
  environments},
\newblock \bibinfo{journal}{Sensors} \bibinfo{volume}{20}
  (\bibinfo{year}{2020}) \bibinfo{pages}{1880}.
\bibitem[{Alajlan et~al.(2013)Alajlan, Koubaa, Chaari, Bennaceur, and
  Ammar}]{alajlan2013global}
\bibinfo{author}{M.~Alajlan}, \bibinfo{author}{A.~Koubaa},
  \bibinfo{author}{I.~Chaari}, \bibinfo{author}{H.~Bennaceur},
  \bibinfo{author}{A.~Ammar},
\newblock \bibinfo{title}{Global path planning for mobile robots in large-scale
  grid environments using genetic algorithms},
\newblock in: \bibinfo{booktitle}{2013 International Conference on Individual
  and Collective Behaviors in Robotics (ICBR)}, \bibinfo{organization}{IEEE},
  \bibinfo{year}{2013}, pp. \bibinfo{pages}{1--8}.
\bibitem[{Phung and Ha(2021)}]{phung2021safety}
\bibinfo{author}{M.~D. Phung}, \bibinfo{author}{Q.~P. Ha},
\newblock \bibinfo{title}{Safety-enhanced uav path planning with spherical
  vector-based particle swarm optimization},
\newblock \bibinfo{journal}{Applied Soft Computing} \bibinfo{volume}{107}
  (\bibinfo{year}{2021}) \bibinfo{pages}{107376}.
\bibitem[{Huang et~al.(2023)Huang, Zhou, Ran, Wang, Chen, and
  Deng}]{huang2023adaptive}
\bibinfo{author}{C.~Huang}, \bibinfo{author}{X.~Zhou},
  \bibinfo{author}{X.~Ran}, \bibinfo{author}{J.~Wang},
  \bibinfo{author}{H.~Chen}, \bibinfo{author}{W.~Deng},
\newblock \bibinfo{title}{Adaptive cylinder vector particle swarm optimization
  with differential evolution for uav path planning},
\newblock \bibinfo{journal}{Engineering Applications of Artificial
  Intelligence} \bibinfo{volume}{121} (\bibinfo{year}{2023})
  \bibinfo{pages}{105942}.
\bibitem[{Ammar et~al.(2016)Ammar, Bennaceur, Ch{\^a}ari, Koub{\^a}a, and
  Alajlan}]{ammar2016relaxed}
\bibinfo{author}{A.~Ammar}, \bibinfo{author}{H.~Bennaceur},
  \bibinfo{author}{I.~Ch{\^a}ari}, \bibinfo{author}{A.~Koub{\^a}a},
  \bibinfo{author}{M.~Alajlan},
\newblock \bibinfo{title}{Relaxed dijkstra and {A*} with linear complexity for
  robot path planning problems in large-scale grid environments},
\newblock \bibinfo{journal}{Soft Computing} \bibinfo{volume}{20}
  (\bibinfo{year}{2016}) \bibinfo{pages}{4149--4171}.
\bibitem[{Panda and Mishra(2018)}]{panda2018survey}
\bibinfo{author}{M.~Panda}, \bibinfo{author}{A.~Mishra},
\newblock \bibinfo{title}{A survey of shortest-path algorithms},
\newblock \bibinfo{journal}{International Journal of Applied Engineering
  Research} \bibinfo{volume}{13} (\bibinfo{year}{2018})
  \bibinfo{pages}{6817--6820}.
\bibitem[{Lee(1961)}]{lee1961algorithm}
\bibinfo{author}{C.~Y. Lee},
\newblock \bibinfo{title}{An algorithm for path connections and its
  applications},
\newblock \bibinfo{journal}{IRE transactions on electronic computers}
  (\bibinfo{year}{1961}) \bibinfo{pages}{346--365}.
\bibitem[{Rubin(1974)}]{rubin1974lee}
\bibinfo{author}{F.~Rubin},
\newblock \bibinfo{title}{The lee path connection algorithm},
\newblock \bibinfo{journal}{IEEE Transactions on computers}
  \bibinfo{volume}{100} (\bibinfo{year}{1974}) \bibinfo{pages}{907--914}.
\bibitem[{Harabor and Grastien(2011)}]{harabor2011online}
\bibinfo{author}{D.~Harabor}, \bibinfo{author}{A.~Grastien},
\newblock \bibinfo{title}{Online graph pruning for pathfinding on grid maps},
\newblock in: \bibinfo{booktitle}{Proceedings of the AAAI Conference on
  Artificial Intelligence}, volume~\bibinfo{volume}{25}, \bibinfo{year}{2011},
  pp. \bibinfo{pages}{1114--1119}.
\bibitem[{Hadlock(1977)}]{hadlock1977shortest}
\bibinfo{author}{F.~Hadlock},
\newblock \bibinfo{title}{A shortest path algorithm for grid graphs},
\newblock \bibinfo{journal}{Networks} \bibinfo{volume}{7}
  (\bibinfo{year}{1977}) \bibinfo{pages}{323--334}.
\bibitem[{Pohl(1970)}]{pohl1970first}
\bibinfo{author}{I.~Pohl},
\newblock \bibinfo{title}{First results on the effect of error in heuristic
  search},
\newblock \bibinfo{journal}{Machine Intelligence} \bibinfo{volume}{5}
  (\bibinfo{year}{1970}) \bibinfo{pages}{219--236}.
\bibitem[{Pearl(1984)}]{pearl1984heuristics}
\bibinfo{author}{J.~Pearl}, \bibinfo{title}{Heuristics: intelligent search
  strategies for computer problem solving}, \bibinfo{publisher}{Addison-Wesley
  Longman Publishing Co., Inc.}, \bibinfo{year}{1984}.
\bibitem[{Pohl(1973)}]{pohl1973avoidance}
\bibinfo{author}{I.~Pohl},
\newblock \bibinfo{title}{The avoidance of (relative) catastrophe, heuristic
  competence, genuine dynamic weighting and computational issues in heuristic
  problem solving},
\newblock in: \bibinfo{booktitle}{Proceedings of the 3rd international joint
  conference on Artificial intelligence}, \bibinfo{year}{1973}, pp.
  \bibinfo{pages}{12--17}.
\bibitem[{K{\"o}ll and Kaindl(1992)}]{koll1992new}
\bibinfo{author}{A.~L. K{\"o}ll}, \bibinfo{author}{H.~Kaindl},
\newblock \bibinfo{title}{A new approach to dynamic weighting},
\newblock in: \bibinfo{booktitle}{Proceedings of the 10th European conference
  on Artificial intelligence}, \bibinfo{year}{1992}, pp.
  \bibinfo{pages}{16--17}.
\bibitem[{Zheng et~al.(1996)Zheng, Lim, and Iyengar}]{zheng1996finding}
\bibinfo{author}{S.-Q. Zheng}, \bibinfo{author}{J.~S. Lim},
  \bibinfo{author}{S.~S. Iyengar},
\newblock \bibinfo{title}{Finding obstacle-avoiding shortest paths using
  implicit connection graphs},
\newblock \bibinfo{journal}{IEEE Transactions on Computer-Aided Design of
  Integrated Circuits and Systems} \bibinfo{volume}{15} (\bibinfo{year}{1996})
  \bibinfo{pages}{103--110}.
\bibitem[{Sturtevant(2012)}]{sturtevant2012benchmarks}
\bibinfo{author}{N.~R. Sturtevant},
\newblock \bibinfo{title}{Benchmarks for grid-based pathfinding},
\newblock \bibinfo{journal}{IEEE Transactions on Computational Intelligence and
  AI in Games} \bibinfo{volume}{4} (\bibinfo{year}{2012})
  \bibinfo{pages}{144--148}.

\end{thebibliography}

\end{document}